\documentclass[letterpaper,10pt]{article} 
\usepackage{opticameet3} %% use version 3 for proper copyright statement

\newcommand\authormark[1]{\textsuperscript{#1}}

\usepackage{amsmath,amssymb}
\usepackage[colorlinks=true,bookmarks=false,citecolor=blue,urlcolor=blue]{hyperref} %pdflatex

\begin{document}

\title{Data-efficient Modeling of Optical Matrix Multipliers Using Transfer Learning}

\author{A. Cem\authormark{1,*}, O. Jovanovic\authormark{1}, S. Yan\authormark{2}, Y. Ding\authormark{1}, D. Zibar\authormark{1}, F. Da Ros\authormark{1}}

\address{\authormark{1}DTU Electro, Technical University of Denmark, DK-2800, Kongens Lyngby, Denmark\\
\authormark{2}School of Optical \& Electrical Information, Huazhong Univ. of Science and Technology, 430074, Wuhan, China}

\email{\authormark{*}alice@dtu.dk} %% email address is required

%\copyrightyear{2022}

\vspace{-.1cm}

% 35 words
\begin{abstract}
We demonstrate transfer learning-assisted neural network models for optical matrix multipliers with scarce measurement data. Our approach uses $<10\%$ of experimental data needed for best performance and outperforms analytical models for a Mach-Zehnder interferometer mesh.
\end{abstract}
\vspace{-.10cm}
\section{Introduction}
\vspace{-.15cm}
Optical neural networks (NNs) have emerged as a high-speed and energy-efficient solution for accelerating machine learning tasks \cite{shastri2021}. Various photonic integrated circuit (PIC) architectures have been proposed for implementing linear layers through optical matrix multiplication (OMM). Specifically, performing unitary transformations using Mach-Zehnder interferometer (MZI) meshes has drawn large attention in recent years \cite{shastri2021,shen2017}.

For OMM, MZI meshes are programmed by tuning the phase shifters associated with the individual MZIs such that a desired weight matrix is realized. A common choice is to use thermo-optic phase shifters, for which analytical models relating the heater voltages to the phase shifts exist \cite{milanizadeh2019}. Programming a chip accurately using such models may be challenging due to fabrication errors and thermal crosstalk, which has led to the emergence of a variety of offline and online calibration techniques \cite{milanizadeh2020,zhang2021}. One such offline technique proposes to model the MZI mesh using a NN that can be trained using experimental measurements. However, this approach requires significantly more measurements compared to analytical models to outperform them \cite{cemjlt}.

In this work, we propose the use of a simple analytical model efitted to the PIC to numerically generate a synthetic dataset, which is used to pre-train a NN model. Then, we apply transfer learning (TL) by re-training a part of the NN model with few experimental measurements to reduce the impact of the discrepancies between the experimental and numerical training data, similar to the proposal of \cite{demoura2022} for Raman amplifiers. The TL-assisted model achieves a root-mean-square modeling error of 1.6 dB for a fabricated PIC, 0.5 dB lower than the analytical model, and it approaches the performance of the NN over the full dataset with only $<10\%$ of experimental data.

\vspace{-.2cm}
\section{Data-efficient Modeling of MZI Meshes with Transfer Learning}
\vspace{-.05cm}
The model for a MZI mesh implementing OMM relates the tunable heater voltages $\mathbf{V}$ to the matrix describing the implemented linear transformation $\mathbf{W}$. An analytical model (AM) that can be trained to model fabrication errors and also account for thermal crosstalk is given below \cite{cemjlt}:
\vspace{-.2cm}
\begin{equation}
W_{i,j} = \alpha_{i,j} \prod_{m \in M_{i,j}} \frac{1}{4} \left| \frac{\sqrt{ER} - 1}{\sqrt{ER} + 1} \pm e^{i (\phi^{(0)}_{m} + \sum_{n=1}^{N_{MZI}} \phi^{(2)}_{m,n} V_n^2)} \right|^2 ,
\vspace{-.2cm}
\label{eqn_model1}
\end{equation}
where $\alpha$ is the optical loss, $M_{i,j}$ is the set of MZIs from input $j$ to output $i$, $ER$ is the MZI extinction ratio, $N_{MZI}$ is the number of MZIs, and $\phi^{(0)}\&\phi^{(2)}$ are the phase parameters to be trained. Alternatively, a NN can model a PIC with higher accuracy compared to AM. However, training a NN model requires a high number of experimental measurements and AM outperforms the NN model when only few are available \cite{cemjlt}. When working with a limited number of measurements, we propose to combine the two modeling approaches: (i) train AM with the available experimental data, (ii) generate synthetic data numerically using AM, (iii) pre-train NN model using synthetic data, (iv) re-train NN model using experimental data to improve accuracy. The NN weights and biases in the first layer are kept constant after pre-training to preserve the knowledge gained from AM, as shown in Fig. \ref{fig1}a).
\vspace{-.2cm}
\section{Experimental Setup and Results}
\begin{figure}[htbp]
\centering
\includegraphics[width=0.95\textwidth]{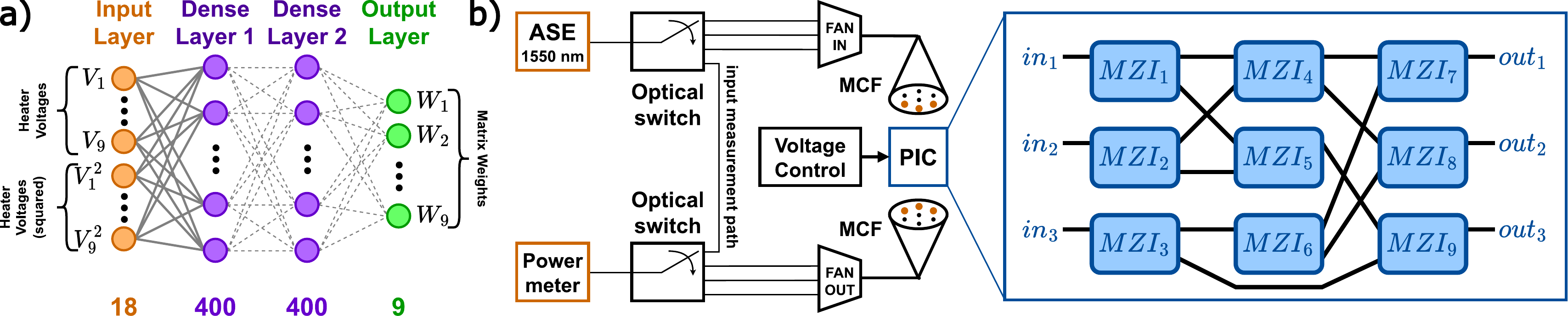}
\caption{(a) Architecture of the TL-assisted NN model for the PIC. Solid lines represent the weights that are fixed during re-training. (b) Experimental measurement setup for data acquisition from the PIC. ASE: amplified spontaneous emission, MCF: multi-core fiber.}
\vspace{-.8cm}
\label{fig1}
\end{figure}

\vspace{-.05cm}
The experimental setup for OMM by a $3\times 3$ matrix is shown in Fig. \ref{fig1}b). Details regarding the PIC and the measurement procedure are discussed in \cite{ding2016} and \cite{cemjlt}, respectively. Values for the $9$ applied voltages were sampled from $9$ i.i.d. uniform distributions from $0$ to $2$ V, which corresponds to one half-period of the MZIs. $400$ sets of $\{\mathbf{V}, \mathbf{W}\}$ were used to train AM and two NN models with and without TL, while $700$ measurements were reserved for testing. A third NN model without TL was trained using $4400$ training measurements for comparison.

\begin{figure}[htbp]
\centering
\includegraphics[width=0.80\textwidth]{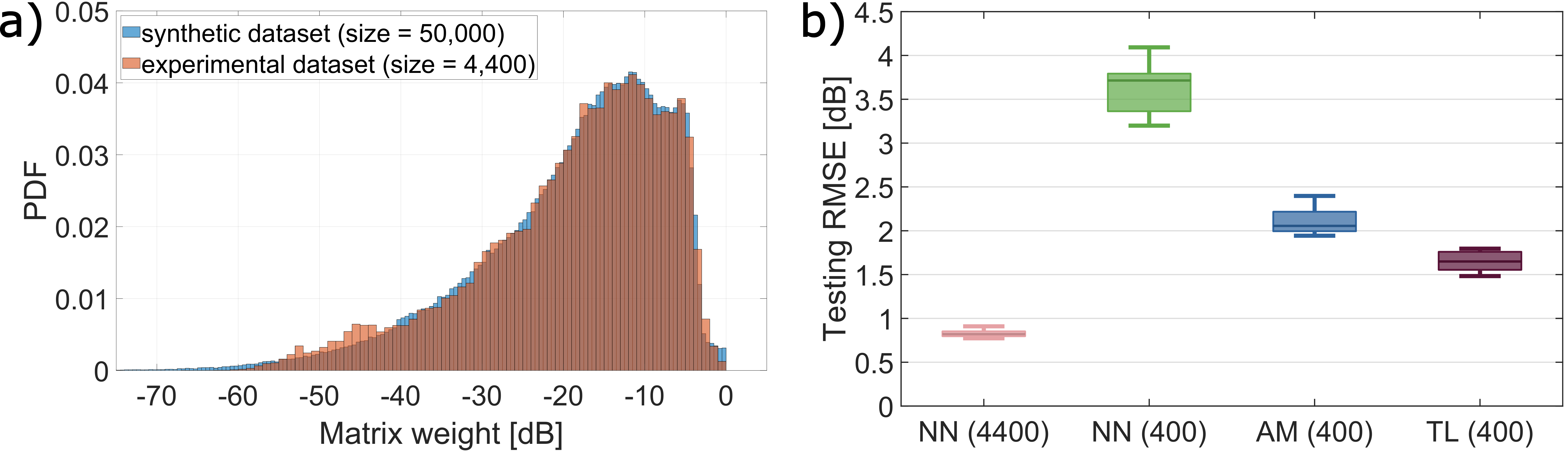}
\caption{(a) Histograms of matrix weights for the synthetic and experimental training datasets, normalized individually to match the estimated probability density functions (PDFs). (b) Testing RMSEs for the models, number of experimental measurements used is given in parentheses. Boxes show $25^{th}$ and $75^{th}$ percentiles while the whiskers show $10^{th}$ and $90^{th}$ percentiles for 20 random seeds.}
\label{fig2}
\vspace{-.6cm}
\end{figure}

AM was trained in MATLAB and was used to generate $50,000$ new synthetic measurements using random voltages as inputs. The histograms for the experimental and synthetic datasets are shown in Fig. \ref{fig2}a). ~ The ~  distributions of weights are very similar for the datasets except for the matrix weights below $-60$ dB, which are only present in the synthetic data. Such datapoints were discarded, resulting in a $<2\%$ reduction in synthetic dataset size. The remaining training dataset was used to pre-train the NN shown in Fig. \ref{fig1}a) with a hyperbolic tangent activation function and $L_1$ and $L_2$ regularization parameters $\lambda_{L1}=5\times 10^{-4}$ and $\lambda_{L2}=9\times 10^{-9}$ using PyTorch with the L-BFGS optimizer. The number of nodes in the hidden layers, $\lambda_{L1}$, and $\lambda_{L2}$ were all optimized on a validation set separately for all 3 NN models to minimize the root-mean-square error (RMSE) between the predicted and measured matrix weights in dB.

The testing RMSEs for the models are shown in Fig. \ref{fig2}b). 20 different seeds were used to randomly obtain 400 samples from the 4400 available experimental measurements as well as initializing the NNs. The results show that while the NN model is able to achieve RMSE $<1$ dB when trained using the entire dataset, it cannot model the PIC with RMSE $<3$ dB when a limited amount of data is available. In contrast, the median RMSEs are $2.1$ and $1.6$ dB for AM and the TL-assisted model, respectively. The NN model obtained using TL clearly outperforms AM for all 20 different training sets with 400 measurements and approaches the performance of the NN model without TL over the full dataset, but requires only $<10\%$ of the experimentally measured data.

\vspace{-.3cm}

\section{Conclusion}

\vspace{-.05cm}

We describe and experimentally evaluate the use of transfer learning to fine-tune a NN model for an optical matrix multiplier by pre-training the NN model using synthetic data generated using a less accurate analytical model. Our proposed approach results in less prediction error compared to using an analytical model or a NN model individually when measurement data is scarce. Transfer learning-assisted NNs can be used to alleviate the practical limitations of data-driven PIC models regarding experimental data acquisition, which is especially critical for larger and more complex MZI mesh architectures.

\small{\noindent\textbf{Acknowledgment} Villum Foundations, Villum YI, OPTIC-AI (no. 29344), ERC CoG FRECOM (no. 771878), National Natural Science Foundation of China (no. 62205114), the Key R\&D Program of Hubei Province (no. 2022BAA001).}


\vspace{-.2cm}

\begin{thebibliography}{99} %% use BibTeX or add references manually

\bibitem{shastri2021} B. Shastri et al., ``Photonics for artificial intelligence and neuromorphic computing," Nat. Phot. 15, 102-114 (2021).

\bibitem{shen2017} Y. Shen et al., ``Deep learning with coherent nanophotonic circuits," Nat. Phot. 11, 441-446 (2017).

\bibitem{milanizadeh2019} M. Milanizadeh et al., “Canceling thermal cross-talk effects in photonic integrated circuits,” JLT 37, 1325-1332, (2019).

\bibitem{milanizadeh2020} M. Milanizadeh et al., ``Control and Calibration Recipes for Photonic Integrated Circuits," JSTQE 26, 1-10 (2020).

\bibitem{zhang2021} H. Zhang et al., ``Efficient On-Chip Training of Optical NNs Using Genetic Algorithm," ACS Photonics 8, (2021).

\bibitem{cemjlt} A. Cem et al., ``Data-driven Modeling of MZI-based Optical Matrix Multipliers," arXiv:2210.09171, (2022).

\bibitem{demoura2022} U. C. de Moura et al., ``Fiber-agnostic machine learning-based Raman amplifier models," JLT, (2022).

\bibitem{ding2016} Y. Ding et al., ``Reconfigurable SDM Switching Using Novel Silicon Photonic Integrated Circuit," Sci. Rep. 6, (2016).

\end{thebibliography}
\end{document}